\documentclass[10pt,twocolumn,letterpaper]{article}

\usepackage{iccv}              

\usepackage{wrapfig}
\usepackage{multirow}
\usepackage{microtype}

\definecolor{iccvblue}{rgb}{0.21,0.49,0.74}
\usepackage[pagebackref,breaklinks,colorlinks,allcolors=iccvblue]{hyperref}
\usepackage[accsupp]{axessibility}
\usepackage{booktabs}  
\usepackage{multirow}  
\usepackage{array}
\usepackage{amsmath}
\usepackage{amsfonts}
\usepackage{xcolor}       
\usepackage{stackengine} 
\usepackage{makecell} 
\usepackage[most]{tcolorbox}
\usepackage{mdframed}
\usepackage{longtable}
\usepackage{tabularx}
\usepackage{adjustbox}
\usepackage{placeins}
\usepackage{float}
\usepackage{cuted} 

\def\paperID{1345} 
\def\confName{ICCV}
\def\confYear{2025}

\title{EMD: Explicit Motion Modeling for High-Quality Street Gaussian Splatting}

\author{
    {\normalsize Xiaobao Wei$^{1,2,3,4,*,\dagger}$ \quad Qingpo Wuwu$^{1,2,*,\dagger}$ \quad Zhongyu Zhao$^{1,2,\dagger}$ \quad Zhuangzhe Wu$^{1}$}\\
    {\normalsize Nan Huang$^{1}$ \quad Ming Lu$^{1,5}$ \quad Ningning Ma$^{2}$ \quad Shanghang Zhang$^{1,\ddagger}$}\\
    {\normalsize $^{1}$State Key Laboratory of Multimedia Information Processing, School of Computer Science, Peking University}\\ {\normalsize $^{2}$Autonomous Driving Development, NIO \quad $^{3}$Institute of Software, Chinese Academy of Sciences}\\
    {\normalsize $^{4}$University of Chinese Academy of Sciences \quad $^{5}$Intel Labs China}\\
    {\normalsize weixiaobao0210@gmail.com}
}

\begin{document}
\maketitle

\begin{strip}
    \vspace{-1.2cm}  
    \centering
    \includegraphics[width=\textwidth]{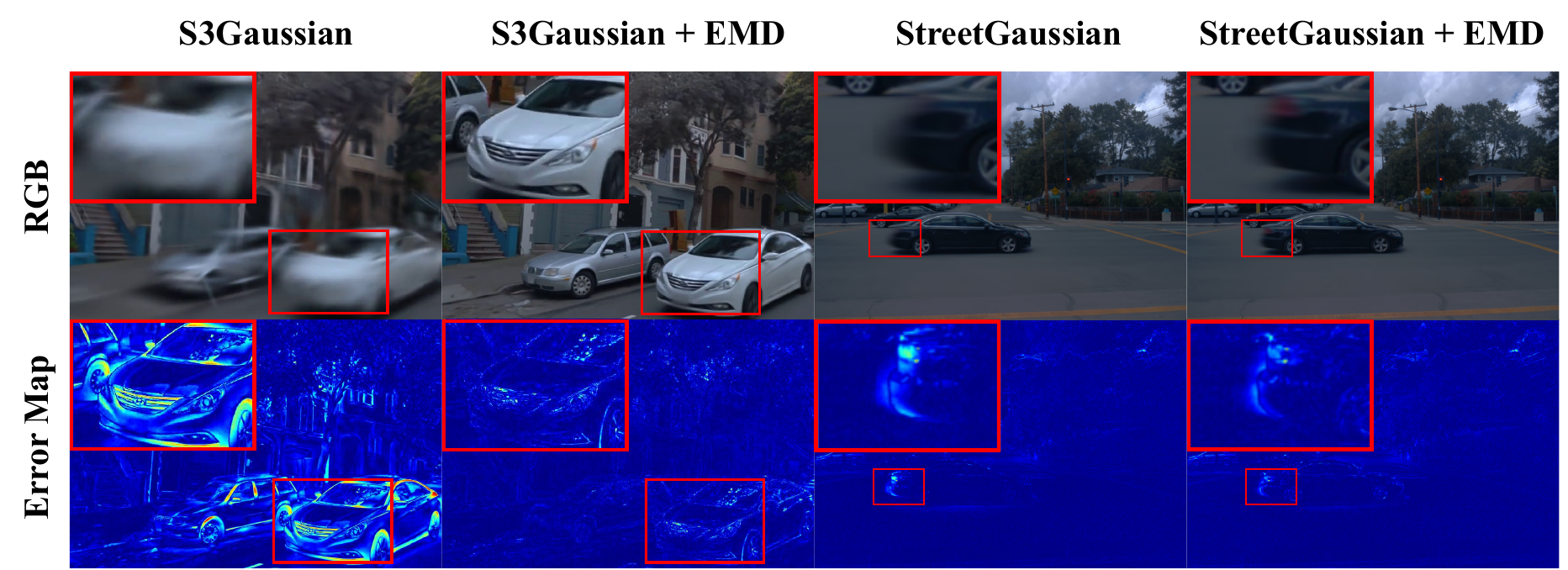}
    \vspace{-7mm}
    \captionsetup{type=figure}
    \captionof{figure}{Previous street Gaussian splatting methods find it challenging to accurately model the motion patterns of dynamic objects, which leads to blurry reconstructions. With the introduction of the proposed Explicit Motion Decomposition (EMD), which compensates for the modeling of dynamic object motion, achieving the state-of-the-art reconstruction quality.}
    \label{fig:teaser}
\end{strip}

\renewcommand{\thefootnote}{\fnsymbol{footnote}} 
\footnotetext[1]{Equal contribution.}
\footnotetext[2]{Work done during internship at NIO.}
\footnotetext[3]{Corresponding author.}
\begin{abstract}
Photorealistic reconstruction of street scenes is essential for developing real-world simulators in autonomous driving. While recent methods based on 3D/4D Gaussian Splatting (GS) have demonstrated promising results, they still encounter challenges in complex street scenes due to the unpredictable motion of dynamic objects. Current methods typically decompose street scenes into static and dynamic objects, learning the Gaussians in either a supervised manner (e.g., w/ 3D bounding-box) or a self-supervised manner (e.g., w/o 3D bounding-box). However, these approaches do not effectively model the motions of dynamic objects (e.g., the motion speed of pedestrians is clearly different from that of vehicles), resulting in suboptimal scene decomposition. To address this, we propose Explicit Motion Decomposition (EMD), which models the motions of dynamic objects by introducing learnable motion embeddings to the Gaussians, enhancing the decomposition in street scenes. 
The proposed plug-and-play EMD module compensates for the lack of motion modeling in self-supervised street Gaussian splatting methods. We also introduce tailored training strategies to extend EMD to supervised approaches. Comprehensive experiments demonstrate the effectiveness of our method, achieving state-of-the-art novel view synthesis performance in self-supervised settings.
The code is available at: \href{https://qingpowuwu.github.io/emd}{https://qingpowuwu.github.io/emd}. 

\end{abstract}

\section{Introduction}
\label{sec:intro}



Novel view synthesis for dynamic street scenes is essential in closed-loop autonomous driving.
Recent advances in neural rendering, particularly Neural Radiance Fields (NeRF)~\cite{2020_08_03-NeRF-Representing_Scenes_as_Neural_Radiance_Fields_for_View_Synthesis} and 3D Gaussian Splatting (3DGS)~\cite{2023_8_08-3dgs_for_real_time_radiance_field_rendering}, have emerged as promising scene reconstruction methods. These pioneering approaches excel in capturing complex geometries and appearances through implicit or explicit neural representations and have been extended to dynamic scenes
~\cite{2020_11_25-Nerfies-Deformable_Neural_Radiance_Fields, 2020_11_27-D-NeRF-Neural_Radiance_Fields_for_Dynamic_Scenes, 2020_11_26-Neural_Scene_Flow_Fields_for_Space-Time-View_Synthesis_of_Dynamic_Scenes, 2020_12_22-Non-Rigid_Neural_Radiance_Fields-Reconstruction_and_Novel_View_Synthesis_of_a_Dynamic_Scene_From_Monocular_Video, 2021_6_24-HyperNeRF-A_Higher_Dimensional_Representation_for_Topologically_Varying_Neural_Radiance_Fields, 2022_11_25-Space-Time_Neural_Irradiance_Fields_for_Free-Viewpoint_Video, 2021_10_26-H-NeRF-Neural_Radiance_Fields_for_Rendering_and_Temporal_Reconstruction_of_Humans_in_Motion, 2022_2_17-Fourier_PlenOctrees_for_Dynamic_Radiance_Field_Rendering_in_Real-time, 2022_10_17-DANOs-Differentiable_Physics_Simulation_of_Dynamics-Augmented_Neural_Objects, 2023_1_23-HexPlane-A_Fast_Representation_for_Dynamic_Scenes, 2024_8_27-FPO++-Efficient_Encoding_and_Rendering_of_Dynamic_Neural_Radiance_Fields_by_Analyzing_and_Enhancing_Fourier_PlenOctrees, wei2024nto3d, wei2024graphavatar}. Despite advancements, reconstructing autonomous driving scenes remains difficult due to complex multi-object dynamics, complex environments, and varied motion patterns. 


To tackle this challenge, existing works based on dynamic NeRF and 3DGS frameworks separate autonomous driving scenes into static and dynamic components through two primary paradigms: Supervised methods acquire priors for dynamic objects, 
such as segmentation masks~\cite{2023_4_SAM_segment_anything}, optical flow~\cite{2023_10_29-dynamo_depth}, or 3D bounding boxes from object tracking. Representative works like StreetGaussian~\cite{2024_01_02-street_gaussian-modelling_dynamic_urban_scenes_with_gs} and OminiRe~\cite{chen2024omnire} demonstrate the effectiveness of using supervision signals for appearance modeling of dynamic objects. In contrast, self-supervised methods achieve static-dynamic separation without explicit supervision, as shown by S3Gaussian~\cite{2024_5_30-S3Gaussian-Self_Supervised_Street_Gaussians_for_Autonomous_Driving} and DeSiRe-GS~\cite{peng2024desire}, which leverage inherent motion cues to optimize a 4D street representation.

While both paradigms yield promising results, supervised methods adopt a binary classification of scene elements as either static or dynamic, which overlooks the continuous spectrum of motion inherent in street scenes. 
Self-supervised methods optimize the entire scene holistically but neglect the varied motion speeds among objects—for example, pedestrians generally move much slower than vehicles. Existing supervised methods~\cite{2024_01_02-street_gaussian-modelling_dynamic_urban_scenes_with_gs, chen2024omnire} mitigate dynamic errors by optimizing 3D bounding boxes, yet self-supervised approaches still lack an effective motion modeling mechanism.

\begin{figure*}[t]
\centering
\includegraphics[width=0.9\textwidth]{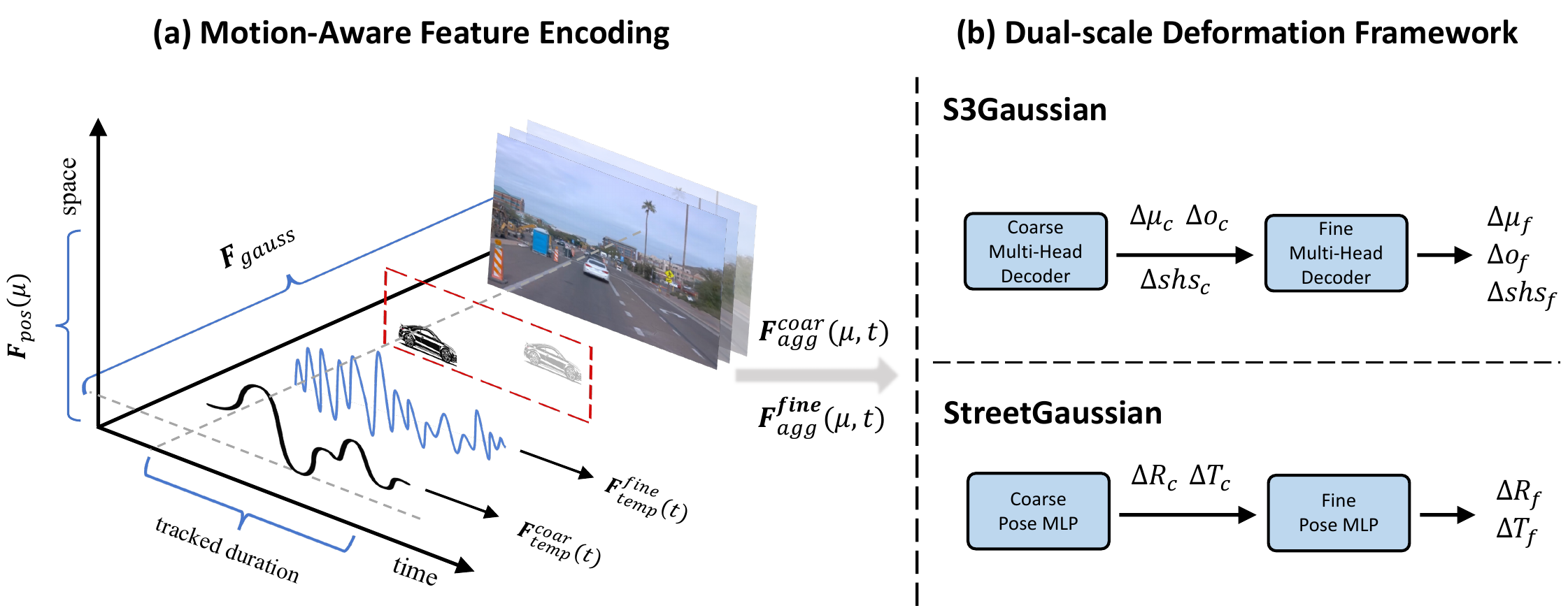}
\vspace{-4mm}
\captionof{figure}{Overview of our Explicit Motion Decomposition (EMD) framework. Given input Gaussian primitives, our method processes them through two main components: (a) Motion-aware Feature Encoding, which combines spatial, temporal, and Gaussian-specific information to capture motion characteristics; and (b) Dual-scale Deformation Framework, which hierarchically models fast global motions and slow local deformations. The framework can be seamlessly integrated into existing supervised and self-supervised approaches.}
\label{fig:method_overview}
\vspace{-6mm}
\end{figure*}

To address the different motion patterns in street scenes, we propose an Explicit Motion Decomposition (EMD) module that can be easily integrated into existing self-supervised frameworks (Fig. \ref{fig:method_overview}). EMD improves scene decomposition by incorporating motion-aware feature encoding and dual-scale deformation modeling. Specifically, we enhance each Gaussian primitive with learnable motion embeddings to capture its motion characteristics and design a hierarchical deformation framework that separately manages fast, global motions and slow, local deformations. This design allows for more efficient analysis of complex street scenes with varying motion speeds. 
We conduct extensive experiments on the Waymo and KITTI datasets by integrating EMD with representative self-supervised methods: S3Gaussian and DeSiRe-GS. In addition, EMD can be seamlessly extended to supervised methods. Our main contributions include:

\begin{itemize}
\item We propose EMD, the first plug-and-play module that effectively addresses varying motion speeds in street scenes through explicit motion modeling. 

\item We introduce tailored training strategies for self-supervised street Gaussian splatting methods and further extend EMD to supervised settings. 

\item Comprehensive experiments on Waymo and KITTI datasets demonstrate previous methods with EMD exhibit better reconstruction quality, achieving state-of-the-art (SOTA) performance in self-supervised settings. 
\end{itemize}

\section{Related Work}

\noindent\textbf{Dynamic Scene Representation. }
Neural Radiance Fields (NeRF)~\cite{2020_08_03-NeRF-Representing_Scenes_as_Neural_Radiance_Fields_for_View_Synthesis} revolutionizes novel view synthesis by introducing volumetric scene representation and has been improved through various extensions~\cite{2021_3_24-Mip-NeRF-A_Multiscale_Representation_for_Anti-Aliasing_Neural_Radiance_Fields, 2021_11_23-Mip-NeRF-360-Unbounded_Anti_Aliased_Neural_Radiance_Fields, 2022_1_16-Instance_NPG-Instant_Neural_Graphics_Primitives_with_a_Multiresolution_Hash_Encoding, 2020_7_2-Neural_Sparse_Voxel_Fields}. However, these methods are restricted to static scenes. 
A common solution involves introducing additional time conditions, as adopted by \cite{2020_11_25-Nerfies-Deformable_Neural_Radiance_Fields, 2020_11_27-D-NeRF-Neural_Radiance_Fields_for_Dynamic_Scenes, 2021_6_24-HyperNeRF-A_Higher_Dimensional_Representation_for_Topologically_Varying_Neural_Radiance_Fields}. 
By introducing additional supervision signals, several works~\cite{2020_5_9-Panoptic_Neural_Fields-A_Semantic_Object-Aware_Neural_Scene_Representation, 2021_03_24-Semantic_NeRF-Neural_Radiance_Fields_for_Semantic_Scene_Segmentation, 2021_4_1-DietNeRF-Putting_NeRF_on_a_Diet_Semantically_Consistent_Few-Shot_View_Synthesis, xu2022-sinnerf-training_neural_radiance_fields_on_complex_scenes, guo2022-nerfren-neural_radiance_fields_with_reflections, deng2022-depth_supervised_nerf-fewer_views_and_faster_training_for_free, niemeyer2022-regnerf-regularizing_neural_radiance_field, 2021_07_20-DepthNeRF-Single_Image_Neural_Radiance_Fields_from_Self_Supervised_Depth, 2022_03_28-Flow_Supervised_Neural_Radiance_Fields_for_Dynamic_Scene_Representation, 2023-DynIBaR-Neural_Dynamic_Image-Based_Rendering} improve the rendering quality in both static and dynamic scenes. 
Despite these advances, NeRF-based methods struggle with long training times and limited ability to handle complex motions. 
Recently, 3D Gaussian Splatting (3DGS)~\cite{2023_8_08-3dgs_for_real_time_radiance_field_rendering} has achieved both high efficiency and superior rendering quality by representing scenes with explicit 3D Gaussian primitives. Several works~\cite{2023_8_18-Dynamic_3D_Gaussians, 2023_10_16-4dgs_other_version-Real-time_Photorealistic_Dynamic_Scene_Representation_and_Rendering_with_4D_Gaussian_Splatting, 2023_9_22-Deformable-3DGS-Deformable_3D_Gaussians_for_High-Fidelity_Monocular_Dynamic_Scene_Reconstruction, 2023_12-7-4dgs-4D_Gaussian_Splatting_for_Real-Time_Dynamic_Scene_Rendering} have extended 3DGS to incorporate temporal information, enabling the 4D scene representation.
These methods lay the groundwork for dynamic scene reconstruction in autonomous driving scenarios.

\noindent\textbf{Autonomous Driving Simulation. }
Traditional autonomous driving simulators like AirSim~\cite{2017_7_18-AirSim-High_Fidelity_Visual_and_Physical_Simulation_for_Autonomous_Vehicles} and CARLA \cite{2017_11_10-CARLA-An_Open_Urban_Driving_Simulator} require extensive manual effort for environment creation while struggling to achieve photorealistic rendering. 
To address this approaches based on neural fields have emerged as promising solutions for simulating street scenes. Early NeRF-based methods~\cite{2020_11_20-NSG-Neural_Scene_Graphs_for_Dynamic_Scenes, 2021_12_20-Mega-NERF_Scalable_Construction_of_Large-Scale_NeRFs_for_Virtual_Fly_Through, 2021_11_29-Urban_Radiance_Fields, 2023_5_2-NFL-Neural_LiDAR_Fields_for_Novel_View_Synthesis} introduce neural scene representation for large-scale urban environments, followed by improvements in efficiency and scalability \cite{2022_2_10-Block-NeRF-Scalable_Large_Scene_Neural_View_Synthesis, 2023_3_25-SUDS-Scalable_Urban_Dynamic_Scenes, 2023_7_20-Urban_Radiance_Field_Representation_with_Deformable_Neural_Mesh_Primitives, 2023_11_9-RealTime_Neural_Rasterization_for_Large_Scenes}. 
For dynamic urban scene reconstruction, several methods~\cite{2020_11_20-NSG-Neural_Scene_Graphs_for_Dynamic_Scenes, 2023_7_27-MARS-An_Instance_aware_Modular_and_Realistic_Simulator_for_Autonomous_Driving, 2023_11_26-NeuRAD-Neural_Rendering_for_Autonomous_Driving, 2024_3_29-Multi-Level_Neural_Scene_Graphs_for_Dynamic_Urban_Environments} introduce supervised scene decomposition using learned scene graphs and latent object representations. 
With the advent of 3DGS, DrivingGaussian~\cite{2024_2_27-drivinggaussian-compusing_gaussian_splatting_for_surronding_dynamic_autonomous_driving_scenes}, StreetGaussian~\cite{2024_01_02-street_gaussian-modelling_dynamic_urban_scenes_with_gs}, HUGS~\cite{2024_3_19-HUGS-Holistic_Urban_3D_Scene_Understanding_via_Gaussian_Splatting} and OmniRe~\cite{chen2024omnire} also adopt this supervised paradigm and develop hierarchical scene representations combining dynamic object graphs with incrementally updated static elements.
To eliminate the need for expensive supervision, 
SUDS~\cite{2023_3_25-SUDS-Scalable_Urban_Dynamic_Scenes} and EmerNeRF~\cite{2023_11_3-EMERNERF-EMERGENT_SPATIAL_TEMPORAL_SCENE_DECOMPOSITION_VIA_SELF-SUPERVISION} leverage the optical flow as decomposition guidance.
PVG~\cite{2024_3_20-PVG-periodic_vibration_gaussian-dynamic_urban_scene_reconstruction}, VDG~\cite{2024_6_26-VDG-Vision-Only_Dynamic_Gaussian_for_Driving_Simulation}, S3Gaussian~\cite{2024_5_30-S3Gaussian-Self_Supervised_Street_Gaussians_for_Autonomous_Driving} and DeSiRe-GS~\cite{peng2024desire} extend the self-supervised setting into 3DGS by a unified representation of both static and dynamic elements. 

However, existing self-supervised methods focus on the scene decomposition, overlooking the diverse motion patterns in street environments. This limitation becomes particularly pronounced when reconstructing objects moving at significantly different speeds. To tackle this challenge, we are the first to explore explicit motion modeling and introduce a plug-and-play Explicit Motion Decomposition (EMD) method. EMD enhances self-supervised approaches by capturing the continuous spectrum of motion, and it can be seamlessly extended to supervised methods.



\section{Preliminaries}
\label{sec:Preliminaries} 
\subsection{3D Gaussian Splatting}
\label{subsec:3dgs}

3D Gaussian Splatting (3DGS)~\cite{2023_8_08-3dgs_for_real_time_radiance_field_rendering} proposes an explicit rendering approach to represent 3D scenes through a collection of 3D Gaussian primitives $\mathbb{G}=\left\{\left(\mu_k, \Sigma_k, \alpha_k, \mathbf{c}_k\right)\right\}_{k=1}^K$, where $K$ is the total number of Gaussians. Each Gaussian primitive represents a probabilistic distribution of density in 3D space, defined by its probability density function:
\vspace{-1mm}
\begin{equation}
G_k(x)=e^{-\frac{1}{2}(x-\mu_k)^T \Sigma_k^{-1}(x-\mu_k)},
\end{equation}
where $x \in \mathbb{R}^3$ represents any point in the world space, $\mu_k \in \mathbb{R}^3$ and $\Sigma_k \in \mathbb{R}^{3 \times 3}$ denote the mean position and covariance matrix in world space, respectively, where the covariance matrix determines the shape and orientation of the Gaussian, $\alpha_k \in [0,1]$ is the opacity, and $\mathbf{c}_k$ encodes the view-dependent color information with spherical harmonics.

For rendering, each 3D Gaussian is projected onto the image plane, where the 3D mean $\mu_k$ is transformed to 2D mean $\mu_k^{2D}$, and the world space covariance matrix $\Sigma_k$ is transformed to screen space as $\Sigma_k^{\prime}=JW\Sigma_k W^TJ^T$, with $W$ and $J$ being the viewing transformation and projective transformation Jacobian matrices. The pixel color at screen space position $x$ is computed through alpha blending:
\vspace{-1mm}
\begin{equation}
C(x)=\sum_{k \in \mathcal{N}(\mathbf{x})} \mathbf{c}_k \alpha_k(x) \prod_{j=1}^{k-1}\left(1-\alpha_j(x)\right),
\end{equation}
where $\alpha_k(x)=\alpha_k \exp \left(-\frac{1}{2}\left(x-\mu_k^{2D}\right)^T {\Sigma_k^{\prime}}^{-1}\left(x-\mu_k^{2D}\right)\right)$ represents the opacity contribution of the $k$-th Gaussian at pixel $x$.
$\mathcal{N}(\mathbf{x})$ represents the set of indices of Gaussians intersecting pixel $x \in \mathbb{R}^2$. In practice, for each Gaussian $k$, we parameterize its covariance matrix $\Sigma_k$ using rotation quaternion $\mathbf{q}_k$ and scaling vector $\mathbf{s}_k$ as:
\vspace{-1mm}
\begin{equation}
\Sigma_k = R(\mathbf{q}_k)S(\mathbf{s}_k)S(\mathbf{s}_k)^TR(\mathbf{q}_k)^T,
\end{equation}
where $R(\mathbf{q}_k)$ is the rotation matrix defined by $\mathbf{q}_k$, and $S(\mathbf{s}_k)$ is the diagonal scaling matrix defined by $\mathbf{s}_k$. 


\subsection{4D Gaussian Splatting}
4D Gaussian Splatting (4DGS)~\cite{2023_12-7-4dgs-4D_Gaussian_Splatting_for_Real-Time_Dynamic_Scene_Rendering, yang2023deformable3dgaussianshighfidelity} extend the static 3DGS framework to handle dynamic scenes by incorporating temporal information. For a dynamic scene captured at different timestamps $t \in [0,T]$, each Gaussian primitive is now characterized by time-varying parameters: $\mathbb{G}(t)=\left\{\left(\mu_k(t), \Sigma_k(t), \alpha_k(t), \mathbf{c}_k(t)\right)\right\}_{k=1}^K$. The probability density function of each Gaussian at time $t$ becomes:
\vspace{-1mm}
\begin{equation}
G_k(x,t)=e^{-\frac{1}{2}(x-\mu_k(t))^T \Sigma_k(t)^{-1}(x-\mu_k(t))},
\end{equation}
Then a deformation field is applied to Gaussians. For each timestamp $t$, the position of each Gaussian is updated as:
\vspace{-1mm}
\begin{equation}
\mu_k(t) = \mu_k(0) + \Delta\mu_k(t),
\end{equation}
where $\mu_k(0)$ is the initial position and $\Delta\mu_k(t)$ is the displacement predicted by a deformation network. Similarly, other parameters such as rotation, scaling, and color can be modeled as temporal offsets from their initial states.


\section{Methodology}
4DGS has shown promising results in dynamic scene reconstruction. However, modeling dynamic street scenes remains challenging due to diverse motion patterns. To better handle the varying motion patterns in street scenes, we propose Explicit Motion Decomposition (EMD), a plug-and-play module that can be seamlessly integrated into existing street Gaussian Splatting methods to enhance their capability in handling dynamic scenarios, as illustrated in Fig.~\ref{fig:method_overview}.

\subsection{Problem Formulation}
\label{subsec:formulation}
Given a set of static 3D Gaussian primitives $\mathbb{G}=\{(\mu_k, \mathbf{s}_k, \mathbf{q}_k, \alpha_k, \mathbf{c}_k)\}_{k=1}^K$ and a timestamp $t$, our goal is to learn a deformation field $\mathcal{D}$ that maps each Gaussian's parameters from their canonical states to their corresponding deformed states at time $t$. For notational simplicity, we omit the Gaussian index $k$ in the following formulation:
\vspace{-2mm}
\begin{equation}
\{\mu_t, \mathbf{s}_t, \mathbf{q}_t, \alpha_t, \mathbf{c}_t\} = \mathcal{D}(\{\mu, \mathbf{s}, \mathbf{q}, \alpha, \mathbf{c}\}, t),
\end{equation}

To effectively handle the diverse motion patterns in street scenes, especially the distinct movements between vehicles and pedestrians, we propose a motion-aware deformation module that processes input Gaussian parameters through two key components: motion-aware feature encoding and dual-scale deformation prediction.

\subsubsection{Motion-aware Feature Encoding}
\label{subsubsec:encoding}
Different types of objects in street scenes exhibit distinct motion characteristics. To capture these varied patterns, we first encode the input Gaussian parameters into a comprehensive feature space that combines spatial, temporal, and Gaussian-specific information:
\begin{equation}
\mathbf{F}_{aggr}(\mu,t) = [\mathbf{F}_{pos}(\mu), \mathbf{F}_{temp}(t), \mathbf{F}_{gauss}],
\end{equation}
The spatial component employs multi-frequency positional encoding:
\begin{equation}
\mathbf{F}_{pos}(\mu) = [\mu, \{\sin(2^i\pi \mu), \cos(2^i\pi \mu)\}_{i=0}^{P-1}],
\end{equation}
where $P=10$ is the number of frequency bands, enabling the network to capture both fine geometric details and global structures. 
For temporal information, we design an adaptive temporal embedding function:
\begin{equation}
\label{eq:temporal_embedding}
\mathbf{F}_{temp}(t) = \mathcal{T}(t, N(i)) = \text{Interp}(\mathbf{W}, t, N(i)),
\end{equation}
where $\mathbf{W} \in \mathbb{R}^{N_{max} \times D}$ is a learnable embedding matrix that captures motion patterns, and $N(i)$ progressively increases from $N_{min}=30$ to $N_{max}=150$ temporal samples during training iteration $i$, allowing the model to gradually capture finer temporal dynamics. \(D=4\) is the temporal embedding dimension. 
During temporal embedding $\mathbf{W}$ optimization, we first downsample $\mathbf{W}$ through bilinear interpolation to create an intermediate embedding matrix of size \(N(i) \times D\) at training iteration $i$, which can be formulated as:
\begin{equation}
\label{eq:temporal_inter}
N(i) = N_{min} + (N_{max} - N_{min}) \cdot \min(i, T) / T,
\end{equation}
where hyperparameter $T=25000$ controls the duration of progressive sampling refinement. Then, for each time stamp t, we perform grid sampling on this intermediate matrix with bilinear interpolation mode to obtain the corresponding temporal embedding vector \(\mathbf{F}_{temp}(t)\).
For Gaussian-specific features, we assign a learnable latent embedding $\mathbf{z}_k \in \mathbb{R}^M$ to each Gaussian $k$, where $\mathbf{F}_{gauss} = \mathbf{z}_k$ and $M=32$, enabling the model to represent individual motion characteristics.

\subsubsection{Dual-scale Deformation Framework}
Given the diverse motion patterns in street scenes, ranging from large vehicular movements to subtle pedestrian motions, we design a hierarchical deformation framework that can effectively handle both scales of motion:
\begin{equation}
\begin{split}
    \mathcal{D}(\mu,t) = &\mathcal{D}_{coarse}(\mathbf{F}_{aggr}(\mu,t)) \\
    &+ \mathcal{D}_{fine}(\mathbf{F}_{aggr}(\mu + \Delta\mu_{coarse},t)),
\end{split}
\end{equation}
The final deformed parameters combine both coarse and fine-scale predictions:
\begin{equation}
\begin{aligned}
\mu_t &= \mu + \Delta\mu_{coarse} + \Delta\mu_{fine} \\
\mathbf{s}_t &= \mathbf{s} + \Delta\mathbf{s}_{coarse} + \Delta\mathbf{s}_{fine} \\
\mathbf{q}_t &= \mathbf{q} \otimes \Delta\mathbf{q}_{coarse} \otimes \Delta\mathbf{q}_{fine},
\end{aligned}
\end{equation}
where $\mathcal{D}_{coarse}$ focuses on modeling large-scale motions such as vehicle translations, while $\mathcal{D}_{fine}$ captures local deformations like articulated movements. Similar to position updates, other Gaussian parameters including opacity $\alpha_t$ and spherical harmonics coefficients $\mathbf{c}_t$ are also updated through this dual-scale framework.

\subsection{Integration with Existing Frameworks}
Current approaches for street scene reconstruction generally fall into supervised and self-supervised paradigms. Having formalized our EMD framework, we now illustrate its integration into representative supervised methods. Then we further extend EMD into supervised approaches. 

\subsubsection{Self-supervised Integration}
For self-supervised street Gaussian splatting methods, we enhance S3Gaussian~\cite{2024_5_30-S3Gaussian-Self_Supervised_Street_Gaussians_for_Autonomous_Driving} and DeSiRe-GS~\cite{peng2024desire} with our motion-aware techniques. 
S3Gaussian originally employs a HexPlane to model the dynamic and static element decomposition and utilizes a Multi-head Gaussian Decoder for deformation prediction. We augment each Gaussian with our learnable embedding $\mathbf{z}_k$ and restructure its decoder into our dual-scale framework. Specifically, both coarse and fine stages predict deformations in position ($\Delta\mu$), opacity ($\Delta\alpha$), and spherical harmonics coefficients ($\Delta\mathbf{c}$), enabling more precise motion modeling through hierarchical refinement. 
As for DeSiRe-GS built upon PVG~\cite{2024_3_20-PVG-periodic_vibration_gaussian-dynamic_urban_scene_reconstruction}, we also assign the learnable Gaussian embedding $\mathbf{z}_k$ into the framework. Then we apply the deformation for the position ($\Delta\mu$) and spherical harmonics coefficients ($\Delta\mathbf{c}$) with the proposed dual-scale deformation framework. We retain the same self-supervised deformation settings as those used in the baseline methods.

\begin{table*}[!ht]
\centering
\caption{S3Gaussian comparison: results on the Waymo Open dataset for scene reconstruction and novel view synthesis.}
\vspace{-3mm}
\label{table1:performance_comparison}
\small
\resizebox{0.9\textwidth}{!}{
\begin{tabular}{c@{\hspace{3pt}}c c c c c c c c c c c}
\toprule
& & \multicolumn{5}{c}{\textbf{Scene Reconstruction}} & \multicolumn{5}{c}{\textbf{Novel View Synthesis}} \\
\cmidrule(lr){3-7} \cmidrule(lr){8-12}
\textbf{Dataset} & \textbf{Methods} & \multicolumn{3}{c}{Full Image} & \multicolumn{2}{c}{Vehicle} & \multicolumn{3}{c}{Full Image} & \multicolumn{2}{c}{Vehicle} \\
\cmidrule(lr){3-5} \cmidrule(lr){6-7} \cmidrule(lr){8-10} \cmidrule(lr){11-12}
& & PSNR↑ & SSIM↑ & LPIPS↓ & PSNR↑ & SSIM↑ & PSNR↑ & SSIM↑ & LPIPS↓ & PSNR↑ & SSIM↑ \\
\midrule
\multirow{5}{*}{Waymo-D32} & EmerNeRF~\cite{2023_11_3-EMERNERF-EMERGENT_SPATIAL_TEMPORAL_SCENE_DECOMPOSITION_VIA_SELF-SUPERVISION} & 28.16 & 0.806 & 0.228 & 24.32 & 0.682 & 25.14 & 0.747 & 0.313 & \textbf{23.49} & 0.660 \\
& 3DGS~\cite{2023_8_08-3dgs_for_real_time_radiance_field_rendering} & 28.47 & 0.876 & 0.136 & 23.26 & 0.716 & 25.14 & 0.813 & 0.165 & 20.48 & \textbf{0.753} \\
& MARS~\cite{2023_7_27-MARS-An_Instance_aware_Modular_and_Realistic_Simulator_for_Autonomous_Driving} & 28.24 & 0.866 & 0.252 & 23.37 & 0.701 & 26.61 & 0.796 & 0.305 & 22.21 & 0.697 \\
& S3Gaussian~\cite{2024_5_30-S3Gaussian-Self_Supervised_Street_Gaussians_for_Autonomous_Driving} & \underline{30.69} & \underline{0.900} & \underline{0.121} & \underline{26.23} & \underline{0.804} & \textbf{26.62} & \underline{0.824} & \underline{0.159} & 22.61 & 0.681 \\
\cmidrule(lr){2-12}
& S3Gaussian+Ours & \textbf{32.50} & \textbf{0.933} & \textbf{0.082} & \textbf{29.04} & \textbf{0.879} & \underline{26.55} & \textbf{0.833} & \textbf{0.126} & \underline{23.39} & \underline{0.703} \\
\bottomrule
\end{tabular}
}
\vspace{-2mm}
\end{table*}

\subsubsection{Supervised Extension}
For supervised settings, though existing methods refine tracked 3D bounding boxes to eliminate rendering errors, we extend EMD into the refinement process, illustrating the need for explicit motion modeling. We select StreetGaussian~\cite{2024_01_02-street_gaussian-modelling_dynamic_urban_scenes_with_gs} and OmniRe~\cite{chen2024omnire} as our baselines. 
StreetGaussian provides a framework that represents dynamic objects through tracked vehicle poses and object-specific Gaussians. Each object is characterized by tracked poses $\{R_t, T_t\}_{t=1}^{N_t}$ that transform object Gaussians from local coordinates ($\mu_o, R_o$) to world coordinates ($\mu_w, R_w$). To enhance its motion modeling capability, we also augment each Gaussian with our learnable embedding $\mathbf{z}_k$ and apply temporal embedding only within each object's tracked duration, effectively capturing object-specific temporal dynamics. Finally, we incorporate our dual-scale framework into the pose optimization:
\vspace{-2mm}
\begin{equation}
\begin{aligned}
R'_t &= \Delta R_t^f \cdot \Delta R_t^c \cdot R_t \\
T'_t &= T_t + (\Delta T_t^c + \Delta T_t^f),
\end{aligned}
\end{equation}
where $\Delta R_t^c, \Delta T_t^c$ handle large pose corrections and $\Delta R_t^f, \Delta T_t^f$ capture subtle adjustments. 
Similarly, we apply the dual-scale framework to appearance modeling, where spherical harmonics coefficients are refined through both coarse and fine stages.

For OmniRe, it further proposes non-rigid SMPL~\cite{loper2023smpl} nodes for human modeling. 
For the rigid nodes, we apply the same tracked box refinement in StreetGaussian into OmniRe. 
We further implement the dual-scale refinement to the SMPL model including pose parameters $\theta_t \in \mathbb{R}^{24 \times 3 \times 3}$ and shape parameters $\beta_t \in \mathbb{R}^{10}$:
\vspace{-2mm}
\begin{equation}
\begin{aligned}
\theta'_t &= \Delta \theta_t^f \cdot \Delta \theta_t^c \cdot \theta_t \\
\beta'_t &= \beta_t + (\Delta \beta_t^c + \Delta \beta_t^f),
\end{aligned}
\end{equation}
For training, in addition to using the same loss function as the baselines, we implement a local smoothness regularization for the learnable Gaussian embeddings. Inspired by ~\cite{luiten2023dynamic3dgaussianstracking}, this regularization encourages neighboring Gaussians $i$ and $j$ to have similar representations:
\begin{equation}
\begin{aligned}
\mathcal{L}_{\mathbf{z}_k}=\frac{1}{d|\mathcal{U}|} \sum_{i\in \mathcal{U}} \sum_{j\in \text{KNN}_{i;d}} (e^{-\lambda_w \|\mu_j-\mu_i\|_2}\| \mathbf{z}_{k_i} - \mathbf{z}_{k_j} \|_2),
\end{aligned}
\end{equation}
where hyperparameters $\lambda_w=2000$ and $d=20$. KNN means the k-nearest-neighbors algorithm. 
We also regularize the predicted coarse and fine deformations, constraining their values to remain close to zero. 
For further training details, please refer to the supplementary materials. 

\begin{table*}[t]
    \centering
    \setlength{\tabcolsep}{2pt}
    \caption{DeSiRe-GS comparison: results on the Waymo Open dataset and KITTI dataset for scene reconstruction and novel view synthesis. }
    \vspace{-3mm}
    \resizebox{0.9\textwidth}{!}{%
    \begin{tabular}{lcccccccccccccc}
        \toprule
       \multirow{3}{*}{\textbf{Method}}  & \multicolumn{7}{c}{\textbf{Waymo Open}} & \multicolumn{7}{c}{\textbf{KITTI}} \\
    \cmidrule(lr){2-8} \cmidrule(lr){9-15} 
         & \multicolumn{3}{c}{{Scene Reconstruction}} & \multicolumn{3}{c}{{Novel View Synthesis}} &  \multirow{2}{*}{{FPS}} & \multicolumn{3}{c}{{Scene Reconstruction}} & \multicolumn{3}{c}{{Novel View Synthesis}} & \multirow{2}{*}{{FPS}} \\
        \cmidrule(lr){2-4} \cmidrule(lr){5-7}  \cmidrule(lr){9-11} \cmidrule(lr){12-14} 
         & PSNR $\uparrow$ & SSIM $\uparrow$ & LPIPS $\downarrow$ & PSNR $\uparrow$ & SSIM $\uparrow$ & LPIPS $\downarrow$ & & PSNR $\uparrow$ & SSIM $\uparrow$ & LPIPS $\downarrow$ & PSNR $\uparrow$ & SSIM $\uparrow$ & LPIPS $\downarrow$ & \\
        \midrule
        S-NeRF~\cite{xie2023s_nerf} & 19.67 & 0.528 & 0.387 & 19.22 & 0.515 & 0.400 & 0.0014 & 19.23 & 0.664 & 0.193 & 18.71 & 0.606 & 0.352 & 0.0075 \\
        StreetSurf~\cite{guo2023streetsurf} & 26.70 & 0.846 & 0.3717 & 23.78 & 0.822 & 0.401 & 0.097 & 24.14 & 0.819 & 0.257 & 22.48 & 0.763 & 0.304 & 0.37 \\
        3DGS~\cite{2023_8_08-3dgs_for_real_time_radiance_field_rendering} & 27.99 & 0.866 & 0.293 & 25.08 & 0.822 & 0.319 & 63 & 21.02 & 0.811 & 0.202 & 19.54 & 0.776 & 0.224 & 125 \\
        NSG~\cite{2020_11_20-NSG-Neural_Scene_Graphs_for_Dynamic_Scenes} & 24.08 & 0.656 & 0.441 & 21.01 & 0.571 & 0.487 & 0.032 & 19.19 & 0.683 & 0.189 & 17.78 & 0.645 & 0.312 & 0.19 \\
        Mars~\cite{2023_7_27-MARS-An_Instance_aware_Modular_and_Realistic_Simulator_for_Autonomous_Driving} & 21.81 & 0.681 & 0.430 & 20.69 & 0.636 & 0.453 & 0.030 & 27.96 & 0.900 & 0.185 & 24.31 & 0.845 & 0.160 & 0.31 \\
        SUDS~\cite{2023_3_25-SUDS-Scalable_Urban_Dynamic_Scenes} & 28.83 & 0.805 & 0.317 & 25.36 & 0.783 & 0.384 & 0.008 & 28.83 & 0.917 & 0.147 & 26.07 & 0.797 & 0.131 & 0.29 \\
        EmerNeRF~\cite{2023_11_3-EMERNERF-EMERGENT_SPATIAL_TEMPORAL_SCENE_DECOMPOSITION_VIA_SELF-SUPERVISION} & 28.11 & 0.786 & 0.373 & 25.92 & 0.763 & 0.384 & 0.053 & 26.95 & 0.828 & 0.218 & 25.24 & 0.801 & 0.237 & 0.28 \\
        PVG~\cite{2024_3_20-PVG-periodic_vibration_gaussian-dynamic_urban_scene_reconstruction} & {32.46} & {0.910} & {0.229} & {28.11} & {0.849} & {0.279} & 50 & {32.83} & {0.937} & {0.070} & {27.43} & {0.896} & {0.114} & 59 \\ 
        DeSiRe-GS~\cite{peng2024desire} & 33.61 & 0.919 & 0.204 & 29.75 & 0.878 & 0.213 & 36 & 33.94 & 0.949 & \textbf{0.040} & 28.87 & 0.901 & 0.106 & 41 \\
        \midrule
        DeSiRe-GS + ours & \textbf{34.15} & \textbf{0.948} & \textbf{0.130} & \textbf{29.91} & \textbf{0.880} & \textbf{0.190} & 32 & \textbf{34.13} & \textbf{0.954} & \textbf{0.040} & \textbf{29.05} & \textbf{0.904} & \textbf{0.094} & 32 \\
        \bottomrule
    \end{tabular}
    }
    \label{tab:desire_performance}
    \vspace{-2mm}
\end{table*}
\begin{figure*}[!ht]
\centering
\includegraphics[width=0.9\textwidth]{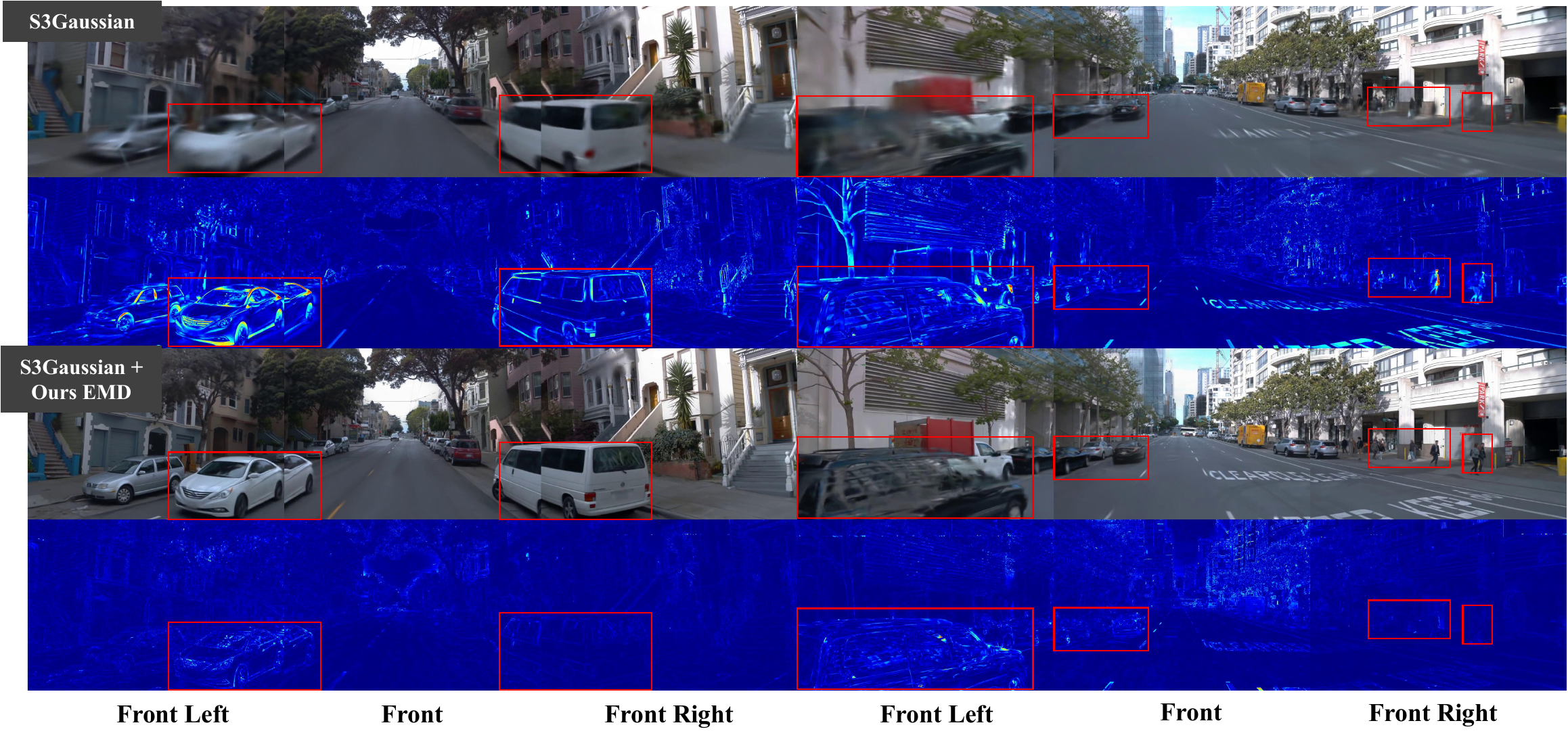}
\vspace{-2mm}
\captionof{figure}{Visualization comparison on the self-supervised setting between S3Gaussian and S3Gaussian+ours EMD. We also visualize the error maps between the rendered images and ground truth to provide further insights. \textbf{Please refer to appendix for more visualization.}}
\label{fig:visualization_s3gaussian}
\vspace{-4mm}
\end{figure*}

\begin{figure*}[!ht]
\centering
\includegraphics[width=0.9\textwidth]{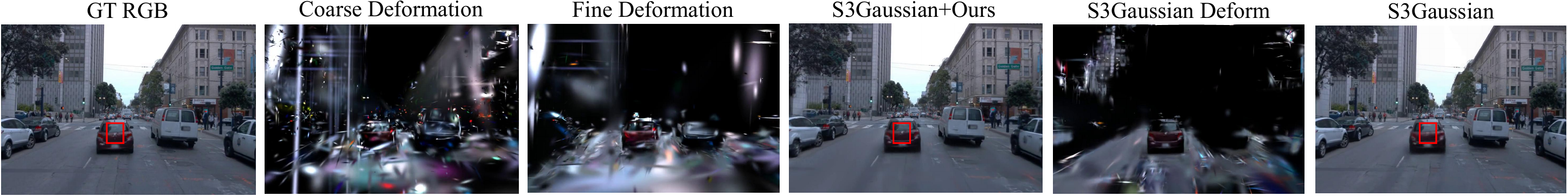}
\vspace{-2mm}
\captionof{figure}{Motion deformation comparison between S3Gaussian + Ours and S3Gaussian. Please zoom in for more details.}
\label{fig4:Motion_comparison}
\vspace{-6mm}
\end{figure*}

\section{Experiments}

\subsection{Datasets and Metrics}
\label{subsec:Datasets and Metrics}
\noindent\textbf{Dataset and Baselines.} We conduct extensive experiments on the Waymo Open Dataset~\cite{sun2020_waymo} and KITTI Dataset~\cite{geiger2012_we_kitti} to comprehensively evaluate our method. For self-supervised setting, we select representative state-of-the-art methods including S3Gaussian~\cite{2024_5_30-S3Gaussian-Self_Supervised_Street_Gaussians_for_Autonomous_Driving} and DeSiRe-GS~\cite{peng2024desire}. To compare with S3Gaussian, we use the dynamic32 (D32) split with 3 frontal cameras of Waymo dataset introduced by EmerNeRF~\cite{2023_11_3-EMERNERF-EMERGENT_SPATIAL_TEMPORAL_SCENE_DECOMPOSITION_VIA_SELF-SUPERVISION} and benchmark S3Gaussian + EMD with vanilla 3DGS~\cite{2023_8_08-3dgs_for_real_time_radiance_field_rendering} and MARS~\cite{2023_7_27-MARS-An_Instance_aware_Modular_and_Realistic_Simulator_for_Autonomous_Driving} on scene reconstruction and novel view synthesis. To compare with DeSiRe-GS which follows PVG~\cite{2024_3_20-PVG-periodic_vibration_gaussian-dynamic_urban_scene_reconstruction}, we use the same subset in PVG, including 4 Waymo scenes and 3 KITTI scenes. We implement baselines including S-NeRF~\cite{xie2023s_nerf}, StreetSurf~\cite{guo2023streetsurf}, NSG~\cite{2021_3_5-Neural_Scene_Graphs_for_Dynamic_Scenes} and SUDS~\cite{2023_3_25-SUDS-Scalable_Urban_Dynamic_Scenes} on scene reconstruction and novel view synthesis. For the supervised setting, we select state-of-the-art approaches including StreetGaussian~\cite{2024_01_02-street_gaussian-modelling_dynamic_urban_scenes_with_gs} and OmniRe~\cite{chen2024omnire}. We select the same Waymo subset from StreetGaussian and OmniRe and conduct experiments on novel view synthesis. It should be noted that StreetGaussian only conducts experiments with one frontal camera in its published version. Thus, we further implement experiments on three frontal cameras for a side-by-side comparison. 

\noindent\textbf{Evaluation Metrics.} 
Following previous protocols, we use PSNR and SSIM to evaluate the pixel-level reconstruction quality. Additionally, we compute LPIPS for perceptual quality assessment. We also report FPS to access inference speed in the comparison between DeSiRe-GS, which refers to frames per second. In addition, we employ the FID metric~\cite{heusel2017gans, zhao2024drivedreamer4d} for novel trajectory synthesis evaluation, which quantifies differences in feature distribution between rendered novel trajectory images and original trajectory images. 

\subsection{Main Results}
\label{subsec:Quantitative Results}
\subsubsection{Self-supervised Performance}

Tab.~\ref{table1:performance_comparison} presents the comparative results on the Waymo-D32 split, where no 3D bounding box annotations were used. 
Our method significantly outperforms previous self-supervised approaches, achieving notable improvements in both full-scene and object-specific metrics. 
These results demonstrate the enhanced ability of our method to accurately model complex motions in street scenes without explicit 3D box annotations. 
S3Gaussian models the entire scene holistically but lacks the ability to capture the diverse motion patterns in dynamic street scenes, whereas EMD effectively compensates for this shortcoming.
For novel view synthesis, our method continues to perform competitively, which reflects its robust generalization to previously unseen viewpoints. 

Tab.~\ref{tab:desire_performance} compares DeSiRe-GS with our proposed EMD. The results show that EMD achieves state-of-the-art performance across various experimental settings. DeSiRe-GS relies on dynamic masks to identify foreground objects, thereby overlooking motion patterns in street scenes. EMD boosts the rendering quality of DeSiRe-GS with only a slight reduction in inference speed. These quantitative results effectively demonstrate that EMD is well-suited for current self-supervised street Gaussian methods.

In addition, we present a side-by-side visualization comparison between S3Gaussian and S3Gaussian+EMD in Fig.~\ref{fig:visualization_s3gaussian}. The error maps, which compare the ground truth to the rendered results, clearly demonstrate that S3Gaussian+EMD outperforms S3Gaussian in modeling dynamic objects with varying motion speeds. 
S3Gaussian implicitly models dynamic vehicles, but it fails to capture changes in speed during motion and the differences in movement between vehicles, leading to blurred reconstructions.
on the contrary, our method captures the distinct motion characteristics of different dynamic objects, leading to more accurate and consistent scene reconstructions. This emphasizes the effectiveness of Explicit Motion Decomposition (EMD) in modeling the motion of dynamic objects, improving the overall decomposition and photorealistic rendering of street scenes. For more visualization videos, \textbf{please watch the webpage Sec.\ref{sec:add_videos} in the supplementary materials.}

We also present a visual comparison of motion between S3Gaussian and S3Gaussian+EMD. As shown in Fig.~\ref{fig4:Motion_comparison}, the proposed dual-scale deformation network generates a coarse deformation to model slower motion and larger-scale geometry and a fine deformation to capture faster motion and finer geometric details in the scene. With the incorporation of EMD, S3Gaussian can capture detailed features of dynamic vehicles, including the car brand logo, as demonstrated in the figure. In contrast, S3Gaussian treats the entire moving car as a single dynamic object, failing to produce clear synthesis results for the dynamic vehicles.

\begin{table}[!ht]
\centering
\setlength\tabcolsep{2pt}
\vspace{-1mm}
\caption{StreetGaussin comparison: novel view synthesis results on the Waymo Open dataset with one and three frontal cameras.}
\vspace{-1mm}
\label{table2:streetgs_comparison}
\small
\begin{tabular}{@{}c@{\hspace{4pt}}c@{\hspace{10pt}}c@{\hspace{10pt}}c@{\hspace{10pt}}c@{}}
\toprule
\multirow{2}{*}[-\dimexpr\ht\strutbox/2]{\textbf{Methods}} & \multicolumn{3}{c}{\textbf{Full Image}} & \textbf{Vehicle} \\
\cmidrule(l{0pt}r{11pt}){2-4} \cmidrule(l{-1pt}r{1pt}){5-5}
& PSNR↑ & SSIM↑ & LPIPS↓ & PSNR↑ \\
\midrule
\multicolumn{5}{c}{\color{gray}{\textit{One camera setting}} }       \\
3DGS~\cite{2023_8_08-3dgs_for_real_time_radiance_field_rendering} & 29.64 & 0.918 & 0.117 & 21.25 \\
NSG~\cite{2020_11_20-NSG-Neural_Scene_Graphs_for_Dynamic_Scenes} & 28.31 & 0.862 & 0.346 & 24.32 \\
MARS~\cite{2023_7_27-MARS-An_Instance_aware_Modular_and_Realistic_Simulator_for_Autonomous_Driving} & 29.75 & 0.886 & 0.264 & 26.54 \\
EmerNeRF~\cite{2023_11_3-EMERNERF-EMERGENT_SPATIAL_TEMPORAL_SCENE_DECOMPOSITION_VIA_SELF-SUPERVISION} & 30.87 & 0.905 & 0.133 & 21.67 \\
StreetGaussian~\cite{2024_01_02-street_gaussian-modelling_dynamic_urban_scenes_with_gs} & 34.61 & 0.938 & 0.079 & 30.23 \\
StreetGaussian + Ours & \textbf{35.41} & \textbf{0.942} & \textbf{0.070} & \textbf{30.96} \\
\midrule
\multicolumn{5}{c}{\color{gray}{\textit{Three camera setting}} }       \\
StreetGaussian & 29.70 & 0.858 & 0.149 & 26.72 \\
StreetGaussian + Ours & \textbf{29.84} & \textbf{0.869} & \textbf{0.145} & \textbf{26.83} \\
\bottomrule
\end{tabular}
\vspace{-6mm}
\end{table}

\begin{table}[ht]
\centering
\caption{OmniRe comparison: novel view synthesis results on the Waymo Open dataset with three frontal cameras.}
\vspace{-1mm}
\normalsize
\resizebox{0.99\linewidth}{!}{%
\begin{tabular}
{ccccccc}
\toprule
\multirow{2}{*}[-\dimexpr\ht\strutbox/2]{\textbf{Methods}} & \multicolumn{2}{c}{\textbf{Full Image}} & \multicolumn{2}{c}{\textbf{Human}} & \multicolumn{2}{c}{\textbf{Vehicle}} \\
\cmidrule(lr){2-3} \cmidrule(lr){4-5} \cmidrule(lr){6-7}
& PSNR↑ & SSIM↑ & PSNR↑ & SSIM↑ & PSNR↑ & SSIM↑ \\
\midrule
EmerNeRF~\cite{2023_11_3-EMERNERF-EMERGENT_SPATIAL_TEMPORAL_SCENE_DECOMPOSITION_VIA_SELF-SUPERVISION} & 29.67 & 0.883 & 20.32 & 0.454 & 22.07 & 0.609 \\
3DGS~\cite{2023_8_08-3dgs_for_real_time_radiance_field_rendering} & 25.57 & 0.906 & 16.62 & 0.387 & 16.00 & 0.407 \\
DeformGS~\cite{yang2023deformable3dgaussianshighfidelity} & 27.72 & 0.922 & 17.30 & 0.426 & 18.91 & 0.530 \\
PVG~\cite{2024_3_20-PVG-periodic_vibration_gaussian-dynamic_urban_scene_reconstruction} & 30.19 & 0.919 & 21.30 & 0.567 & 22.28 & 0.679 \\
HUGS~\cite{2024_3_19-HUGS-Holistic_Urban_3D_Scene_Understanding_via_Gaussian_Splatting} & 27.65 & 0.914 & 15.99 & 0.378 & 23.27 & 0.748 \\
StreetGS~\cite{2024_01_02-street_gaussian-modelling_dynamic_urban_scenes_with_gs} & 28.54 & 0.928 & 16.55 & 0.393 & 26.71 & 0.846 \\
OmniRe~\cite{chen2024omnire} & 32.57 & 0.942 & 24.36 & 0.727 & 27.57 & 0.858 \\
\midrule
OmniRe + Ours & \textbf{33.89} & \textbf{0.958} & \textbf{25.97} & \textbf{0.742} & \textbf{27.82} & \textbf{0.859} \\
\bottomrule
\end{tabular}
}
\label{tab:omnire_results}
\vspace{-2mm}
\end{table}

\subsubsection{Supervised Performance}
To demonstrate the flexibility of EMD, we further extend it to enhance motion modeling in supervised methods. Table~\ref{table2:streetgs_comparison} and Table~\ref{tab:omnire_results} show novel view synthesis comparisons between StreetGaussian and OmniRe, respectively. While these supervised methods employ 3D box refinement to mitigate tracking errors, the results indicate that EMD serves as a valuable complement to this process. In the OmniRe comparison, EMD also benefits non-rigid objects such as humans, whose motion patterns differ significantly from those of vehicles.
Due to the space limitation, please refer to the supplementary for visualization on the supervised setting.

\begin{table}[!ht]
\centering
\vspace{-2mm}
\caption{Novel trajectory synthesis on Waymo dataset.}
\vspace{-2mm}
\label{tab:lane_change}
\small
\begin{tabular}{@{}c@{\hspace{4pt}}c@{\hspace{10pt}}c@{\hspace{10pt}}c@{}}
\toprule
\multirow{2}{*}[-\dimexpr\ht\strutbox/2]{\textbf{Method}} & \multicolumn{3}{c}{\textbf{FID↓}} \\
\cmidrule(l{0pt}r{0pt}){2-4} 
 & 0.5m & 1.0m & 1.5m \\
\midrule
S3Gaussian & 83.48 & 110.11 & 134.38  \\
S3Gaussian + ours & \textbf{45.11} & \textbf{70.26} & \textbf{90.20}  \\
\bottomrule
\end{tabular}
\vspace{-4mm}
\label{tab:novel_traj}
\end{table}

\begin{figure}[!ht]
    \centering
    \includegraphics[width=1.0\linewidth]{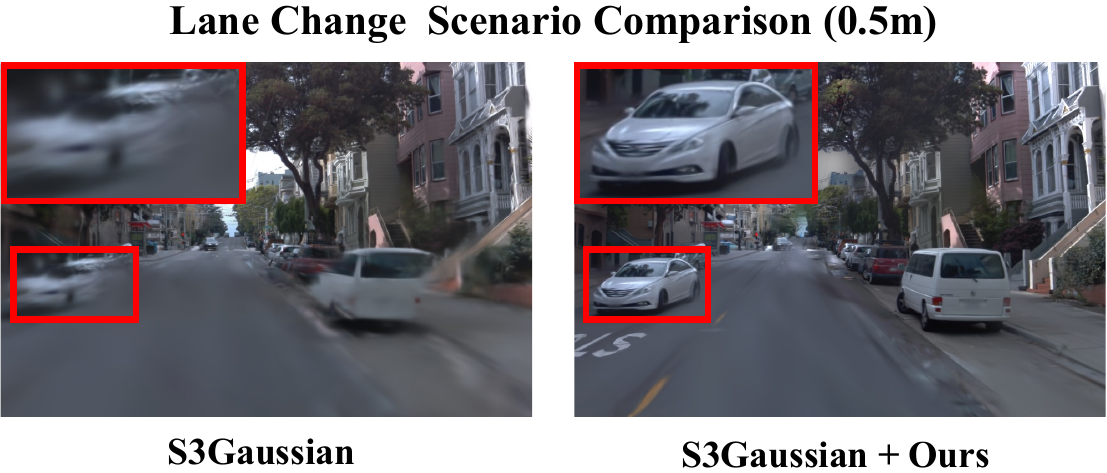}
    \vspace{-3mm}
    \caption{Visualization for novel trajectory synthesis (0.5m offset). \textbf{Please watch the webpage in the supplementary materials for more results.}}
    \label{fig: novel_traj}
    \vspace{-2mm}
\end{figure}

\subsection{Novel Trajectory Synthesis}
Previous novel view synthesis evaluations are limited to interpolated views along the original camera trajectory, which fail to assess the exact simulation performance of the model. Therefore, we compare the FID performance of self-supervised methods and our approach under novel trajectory synthesis. Specifically, we shift the original camera trajectory to the left and right by different offsets (0.5 m, 1.0 m, 1.5 m) and render images along these new trajectories. The results are presented in Tab.~\ref{tab:novel_traj} and one visualization sample is shown in Fig.~\ref{fig: novel_traj}. These findings demonstrate that EMD enhances the accurate modeling of dynamic objects by learning diverse motion patterns, which benefits lane change scenarios. For novel trajectory synthesis videos, \textbf{please refer to the supplementary materials.}

\begin{table}[!ht]
\centering
\caption{Ablation study on the 32 dynamic scenes Waymo subset showing the impact of different components in our framework. All variants are evaluated in self-supervised scene reconstruction. }
\vspace{-2mm}
\label{table3:ablation}
\small
\begin{tabular}{@{}c@{\hspace{4pt}}c@{\hspace{10pt}}c@{\hspace{10pt}}c@{\hspace{10pt}}c@{}}
\toprule
\multirow{2}{*}[-\dimexpr\ht\strutbox/2]{\textbf{Variant}} & \multicolumn{3}{c}{\textbf{Full Image}} & \textbf{Vehicle} \\
\cmidrule(l{0pt}r{11pt}){2-4} \cmidrule(l{-1pt}r{1pt}){5-5}
& PSNR↑ & SSIM↑ & LPIPS↓ & PSNR↑ \\
\midrule
Full Model & \textbf{32.50} & \textbf{0.933} & \textbf{0.082} & \textbf{29.04} \\
w/o Gaussian Embedding & 32.21 & 0.928 & 0.089 & 28.80 \\
w/o Temporal Embedding & 32.23 & 0.922 & 0.091 & 28.08 \\
w/o Coarse Deformation & 29.40 & 0.890 & 0.146 & 24.54 \\
w/o Fine Deformation & 32.45 & 0.931 & 0.118 & 28.80 \\
\bottomrule
\end{tabular}
\vspace{-2mm}
\end{table}

\begin{figure*}[!ht]
    \centering
    \includegraphics[width=0.95\linewidth]{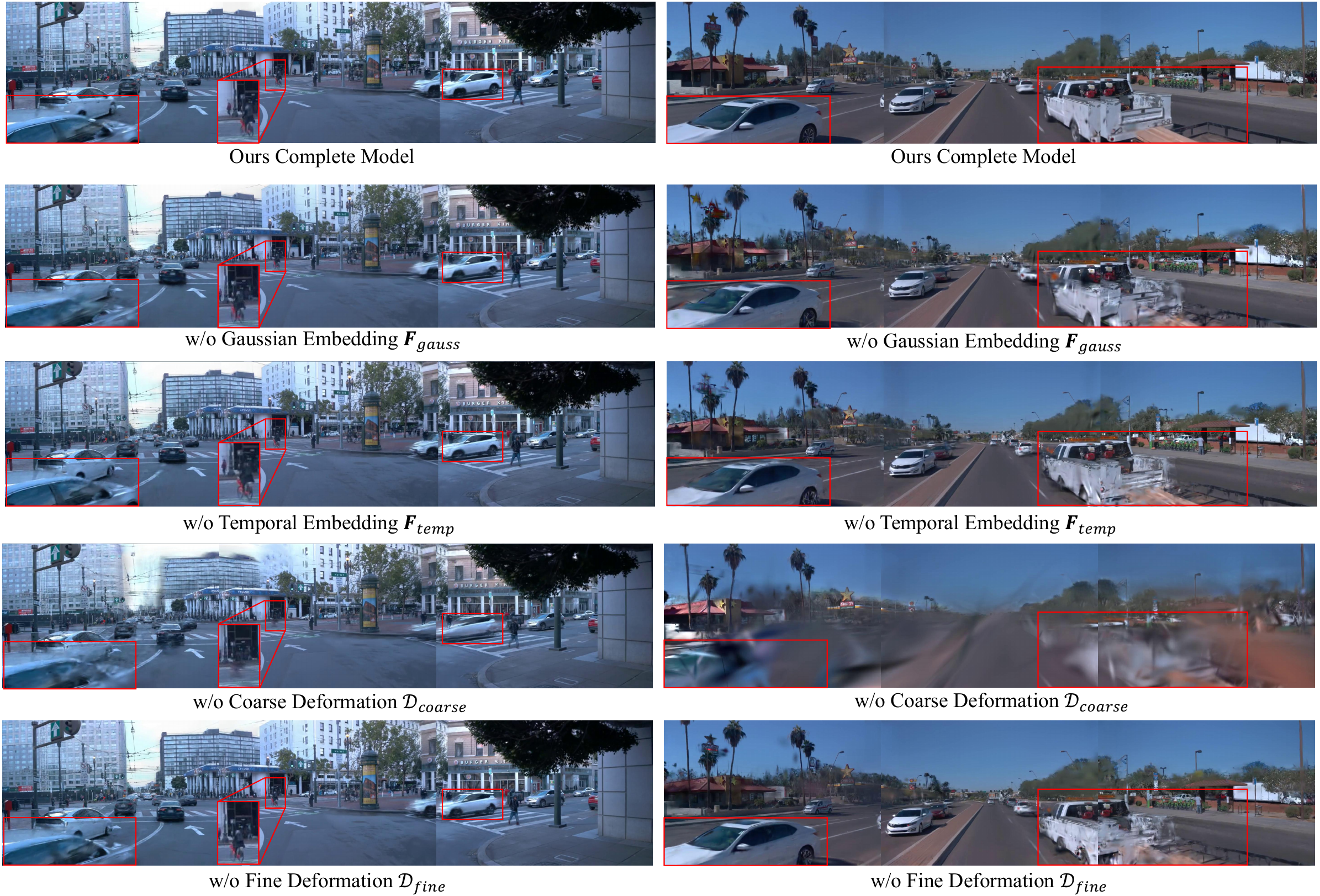}
    \caption{Qualitative ablation study results across three camera views from the Waymo dataset. (a) Our complete model achieves sharp and consistent reconstruction. (b) Removing Gaussian embedding $\mathcal{F}_{gauss}$ leads to blurry object boundaries. (c) Without temporal embedding $\mathcal{F}_{temp}$, results show motion artifacts. (d) Without coarse deformation $\mathcal{D}_{coarse}$, geometric consistency is lost. (e) Absence of fine deformation $\mathcal{D}_{fine}$ causes detail degradation.}
    \label{fig:ablation_vis}
    \vspace{-2mm}
\end{figure*}

\subsection{Ablation Studies}

\label{subsec:Ablation Studies}

To assess the contribution of each component in our framework, we perform comprehensive ablation studies, as shown in Tab.~\ref{table3:ablation} and Fig.~\ref{fig:ablation_vis}. The results reveal the critical role of the Gaussian embedding, as its removal leads to the performance drop, with a reduction of 0.29 PSNR (32.21 vs. 32.50). This indicates that the Gaussian embedding is essential for effectively capturing the motion characteristics for each dynamic gaussian. 
The temporal embedding also plays a crucial role, with its absence leading to a 0.27 PSNR drop (32.23 vs. 32.50), underscoring its importance in modeling the temporal variation of object motion over time.

Both the coarse and fine deformation components are integral to the final performance, with their removal leading to performance degradation. Specifically, excluding the coarse deformation component causes a significant 3.10 PSNR drop (29.40 vs. 32.50), suggesting that the coarse adjustments are vital for maintaining the overall scene structure. On the other hand, removing the fine deformation component results in a 0.036 LPIPS increase (0.118 vs. 0.082), implying that fine deformation refines the details and its absence worsens the perceptual quality. The ablation study demonstrates that our proposed motion-aware feature encoding and dual-scale deformation effectively model dynamic objects with varying motion speeds, which is crucial for enhancing the reconstruction quality of existing street Gaussians.



\vspace{-2mm}
\section{Conclusion}
In this paper, we present EMD, the first plug-and-play module that effectively handles varying motion speeds in street scene reconstruction. 
By introducing motion-aware feature encoding and dual-scale deformation modeling, our approach successfully captures complex motion patterns in real-world scenarios. Comprehensive experiments on the Waymo and KITTI datasets demonstrate that EMD achieves the state-of-the-art novel view synthesis quality for self-supervised frameworks. EMD can be extended into supervised street Gaussian splatting methods, which opens up new possibilities for high-quality driving scene reconstruction. 
\paragraph{Acknowledgments.} This work was supported by the National Science and Technology Major Project (No. 2022ZD0117800). 

{
    \small
    \bibliographystyle{ieeenat_fullname}
    \bibliography{main}
}
\clearpage
\setcounter{page}{1}
\maketitlesupplementary
\appendix

\section{Overview}

The supplementary material includes the subsequent components. 
\begin{itemize}
    \item Additional Visualization Videos
    \item Implementation Details
    \begin{itemize}
        \item[--] Training Schemes
        \item[--] Training Details
        \item[--] Parameters and Efficiency
    \end{itemize}
    \item Parameter Sensitivity
\end{itemize}

\section{Additional Visualization Videos}
\label{sec:add_videos}
Please double-click the ``Demo Webpage-Please wait until loaded.html" file and open it in your browser. This offline webpage contains videos covering the following experiments:
\begin{itemize}
    \item Self-supervided Comparison
    \item Box Supervised Comparison
    \item Novel Trajectory Synthesis
    \item Temporal Embedding Evolution
\end{itemize}
Due to the numerous videos, \textbf{please wait for the webpage until loaded.}

\section{Implementation Details}

\subsection{Training Schemes}

\label{sec/optimize}
\textbf{LiDAR Prior Initialization.}
To initialize the positions of the 3D Gaussians, we leverage the LiDAR point cloud captured by the vehicle instead of using the original SFM~\cite{Schonberger_Frahm_2016} point cloud to provide a better geometric structure. To reduce model size, we also downsample the entire point cloud by voxelizing it and filtering out points outside the image. For colors, we initialize them randomly.

\noindent\textbf{Optimization Objective.} 
Following Street Gaussian, we introduce the sky supervision loss $L_{sky}$ into the original loss function proposed by S3Gaussian. Subsequently, we get a composed training loss function which can impose various constraints to our model.

\begin{equation}
\begin{split}
\mathcal{L} = \mathcal{L}_{color} &+ \lambda_{depth}\mathcal{L}_{depth} + \lambda_{feat}\mathcal{L}_{feat} \\
&+ \lambda_{tv}\mathcal{L}_{tv} + \lambda_{sky}\mathcal{L}_{sky} + \lambda_{reg}\mathcal{L}_{reg}
\end{split}
\end{equation}
Here, $\mathcal{L}_{depth}$ is the mean square error (MSE) loss between the rendered depth map and the estimated depth map from the LiDAR point cloud, which aids in supervising the expected position of 3D Gaussians. $\mathcal{L}_{feat}$ is also the L2 loss of semantic features to reduce the gap between both planes. $\mathcal{L}_{tv}$ is a total-variational loss based on grids to make rendered objects smoother. $\mathcal{L}_{color}$ is the main loss to give constraints to the reconstruction process formulated by:

\begin{equation}
    \mathcal{L}_{color}=\mathcal{L}_{rgb} + \lambda_{ssim}\mathcal{L}_{sim}
\end{equation}

\noindent Furthermore, $\mathcal{L}_{reg}$ is organized as:

\begin{equation}
    \mathcal{L}_{reg}=\mathcal{L}_{z_k}+\mathcal{L}_\Delta
\end{equation}
where $\mathcal{L}_{z_k}$ is local smoothness regularization for Gaussian embeddings in the method section. $\mathcal{L}_\Delta$ represents a combination of regularization for coarse and fine deformations, restricting their values near zero. We also detail the coefficients for loss in Tab.~\ref{tab:loss_coefficients}.
\begin{table}[ht]
\caption{Loss function coefficients}
\label{tab:loss_coefficients}
\centering
\begin{tabular}{llllll}
\toprule
$\lambda_{depth}$ & $\lambda_{feat}$ & $\lambda_{feat}$ & $\lambda_{tv}$ & $\lambda_{sky}$ & $\lambda_{reg}$ \\
\midrule
0.5 & 0.1 & 0.1 & 0.1 & 0.1 & 0.01 \\
\bottomrule
\end{tabular}
\end{table}

\begin{table}[!ht]
\centering
\caption{Parameter sensitivity analysis on the D32 dataset, highlighting the effect of varying the dimensions of Gaussian embeddings $\mathbf{z}_k$ and temporal embeddings $\mathbf{z}_w$. All experiments are conducted in the self-supervised setting. Best performances are highlighted in \textbf{bold}. ↑ indicates higher is better, while ↓ indicates lower is better. We also include the changes in model parameters relative to the adopted setting.}
\vspace{-2mm}
\label{table:parameter}
\small
\begin{tabular}{@{}c@{\hspace{8pt}}c@{\hspace{10pt}}c@{\hspace{10pt}}c@{\hspace{10pt}}c@{\hspace{10pt}}c@{}}
\toprule
\multirow{2}{*}[-\dimexpr\ht\strutbox/2]{$\mathbf{z}_k$/$\mathbf{z}_w$} & \multirow{2}{*}[-\dimexpr\ht\strutbox/2]{\textbf{Parameters}} & \multicolumn{3}{c}{\textbf{Full Image}} & \textbf{Vehicle} \\
\cmidrule(l{0pt}r{11pt}){3-5} \cmidrule(l{-1pt}r{1pt}){6-6}
& & \textbf{PSNR↑} & \textbf{SSIM↑} & \textbf{LPIPS↓} & \textbf{PSNR↑} \\
\midrule
32/4 & / & \textbf{32.50} & \textbf{0.933} & \textbf{0.082} & 29.04 \\
128/4 & +14400 & 32.22 & 0.925 & 0.086 & \textbf{29.05} \\
8/4 & -3600 & 31.25 & 0.910 & 0.128 & 27.75 \\
\midrule
32/4 & / & \textbf{32.50} & \textbf{0.933} & 0.082 & \textbf{29.04} \\
32/16 & +14.42M & 32.38 & 0.930 & \textbf{0.081} & 29.01 \\
32/1 & -3.60M & 30.55 & 0.898 & 0.136 & 27.04 \\
\bottomrule
\end{tabular}
\vspace{-4mm}
\end{table}

\subsection{Training Details}
For S3Gaussian, we train the entire pipeline for 50,000 iterations using the Adam optimizer. Following the original S3Gaussian setup, we perform a warm-up phase for each scene, employing 5,000 iterations to train a coarse representation using vanilla 3D Gaussians. After this warm-up phase, we integrate the proposed dual-scale deformation network, which is jointly optimized with the HexPlane. To implement a coarse-to-fine training strategy, temporal embeddings $N(i)$ are progressively increased from $N_{min}$ to $N_{max}$ in 20,000 iterations, allowing for the gradual motion modeling of objects.
Since S3Gaussian is evaluated on 50 frames per clip for each scene, we ensure a fair comparison by conducting all self-supervised validation experiments on the first clip of 32 dynamic scenes. Other configurations, including the detailed setup of the HexPlane and learning rates, are kept consistent with the S3Gaussian.
For StreetGaussian, the entire method is trained for 30,000 iterations on a subset of eight selected scenes from the StreetGaussian dataset. Unlike the self-supervised method, we bind the proposed EMD to the vehicle Gaussians in each scene. Temporal embeddings are applied based on the time each vehicle appears within the scene.
All other settings, including detailed configurations, remain consistent with those described in StreetGaussian. All experiments are conducted on a single NVIDIA A800 GPU.

\section{Parameter Sensitivity}
To analyze the sensitivity of model performance to the dimensions of Gaussian embeddings $\mathbf{z}_k$ and temporal embeddings $\mathbf{z}_w$ (derived from the learnable embedding matrix $\mathbf{W}$), we conduct experiments by varying these dimensions. In the original setup, $\mathbf{z}_k$ is set to 32 and $\mathbf{z}_w$ to 4. Tab.~\ref{table:parameter} summarizes the results, demonstrating how these changes influence performance under the self-supervised setting.

The results reveal that reducing the embedding dimensions leads to significant performance degradation. This is primarily due to the reduced capacity to effectively model the motion of dynamic objects, which is crucial for high-quality reconstruction. On the other hand, increasing the embedding dimensions offers only marginal performance improvements. However, due to the large number of Gaussians in the driving scenes, the higher embedding dimensions result in a substantial increase in the total number of model parameters, leading to a higher computational cost. These findings highlight the trade-off between embedding dimension size and overall model efficiency. While lower dimensions compromise the ability to capture dynamic motion, higher dimensions introduce considerable overhead without proportional gains in performance. The adopted embedding configuration achieves a good balance, maintaining strong performance while keeping the parameter count manageable.

\section{Limitation}
Although EMD effectively addresses the challenge of modeling dynamic objects with varying speeds by incorporating learnable embeddings, some limitations remain. Existing street Gaussian methods do not account for environmental lighting, yet lighting effects play a crucial role in the quality of reconstructions under different lighting conditions. In future work, we plan to explore the possibility of developing a plug-and-play technique to enhance lighting effects in existing methods.

\end{document}